\documentclass{article}
\usepackage{smc2017}
\usepackage{times}
\usepackage{ifpdf}
\usepackage[english]{babel}
\usepackage{cite}

\def\papertitle{\textit{Live Orchestral Piano}, a system for real-time orchestral music generation}
\def\firstauthor{L\'eopold Crestel}
\def\secondauthor{Philippe Esling}
\def\thirdauthor{Third author}

\usepackage[pdftex,
pdftitle={\papertitle},
pdfauthor={\firstauthor, \secondauthor, \thirdauthor},
bookmarksnumbered, 
pdfstartview=XYZ 
]{hyperref}

\usepackage[pdftex]{graphicx}
\graphicspath{{../Figures/}}
\DeclareGraphicsExtensions{.pdf,.jpeg,.png}

\usepackage[figure,table]{hypcap}

\usepackage{amsmath,amssymb} 
\usepackage{bm}
\usepackage{graphicx} 
\graphicspath{{../Figures/}} 
\usepackage{makecell}

\hypersetup{
    colorlinks,%
    citecolor=black,%
    filecolor=black,%
    linkcolor=black,%
    urlcolor=black
}

\title{\papertitle}

 \twoauthors
   {\firstauthor} {IRCAM - CNRS UMR 9912\\ %
     {\tt \href{mailto:leopold.crestel@ircam.fr}{leopold.crestel@ircam.fr}}}
   {\secondauthor} {IRCAM - CNRS UMR 9912\\ %
     {\tt \href{mailto:philippe.esling@ircam.fr}{philippe.esling@ircam.fr}}}

\begin{document}
\capstartfalse
\maketitle
\capstarttrue
\begin{abstract}
This paper introduces the first system performing \textit{automatic orchestration} from a real-time piano input. We cast this problem as a case of \emph{projective orchestration}, where the goal is to learn the underlying regularities existing between piano scores and their orchestrations by well-known composers, in order to later perform this task automatically on novel piano inputs. To that end, we investigate a class of statistical inference models based on the \textit{Restricted Boltzmann Machine} (\textit{RBM}). We introduce an evaluation framework specific to the projective orchestral generation task that provides a quantitative analysis of different models.  We also show that the frame-level accuracy currently used by most music prediction and generation system is highly biased towards models that simply repeat their last input. As prediction and creation are two widely different endeavors, we discuss other potential biases in evaluating temporal generative models through prediction tasks and their impact on a creative system. Finally, we provide an implementation of the proposed models called \textit{Live Orchestral Piano} (LOP), which allows for anyone to play the orchestra in real-time by simply playing on a MIDI keyboard. To evaluate the quality of the system, orchestrations generated by the different models we investigated can be found on a companion website\footnote{\url{https://qsdfo.github.io/LOP/}}.
%
%
\end{abstract}

\section{Introduction}
\textit{Orchestration} is the subtle art of writing musical pieces for the orchestra, by combining the properties of various instruments in order to achieve a particular sonic rendering \cite{koechli_orch,Rimsky-Korsakov:1873aa}. As it extensively relies on spectral characteristics, orchestration is often referred to as the art of manipulating instrumental timbres \cite{mcadams2013timbre}. \textit{Timbre} is defined as the property which allows listeners to distinguish two sounds produced at the same pitch and intensity.
Hence, the sonic palette offered by the pitch range and intensities of each instrument is augmented by the wide range of expressive timbres produced through the use of different playing styles.
Furthermore, it has been shown that some instrumental mixtures can not be characterized by a simple summation of their spectral components, but can lead to a unique \textit{emerging timbre}, with phenomenon such as the orchestral blend \cite{tardieu2012perception}.
Given the number of different instruments in a symphonic orchestra, their respective range of expressiveness (timbre, pitch and intensity), and the phenomenon of emerging timbre, one can foresee the extensive combinatorial complexity embedded in the process of orchestral composition. This complexity have been a major obstacle towards the construction of a scientific basis for the study of orchestration and it remains an empirical discipline taught through the observation of existing examples \cite{piston-orch}.

\begin{figure}
\centering
\includegraphics[scale=0.15]{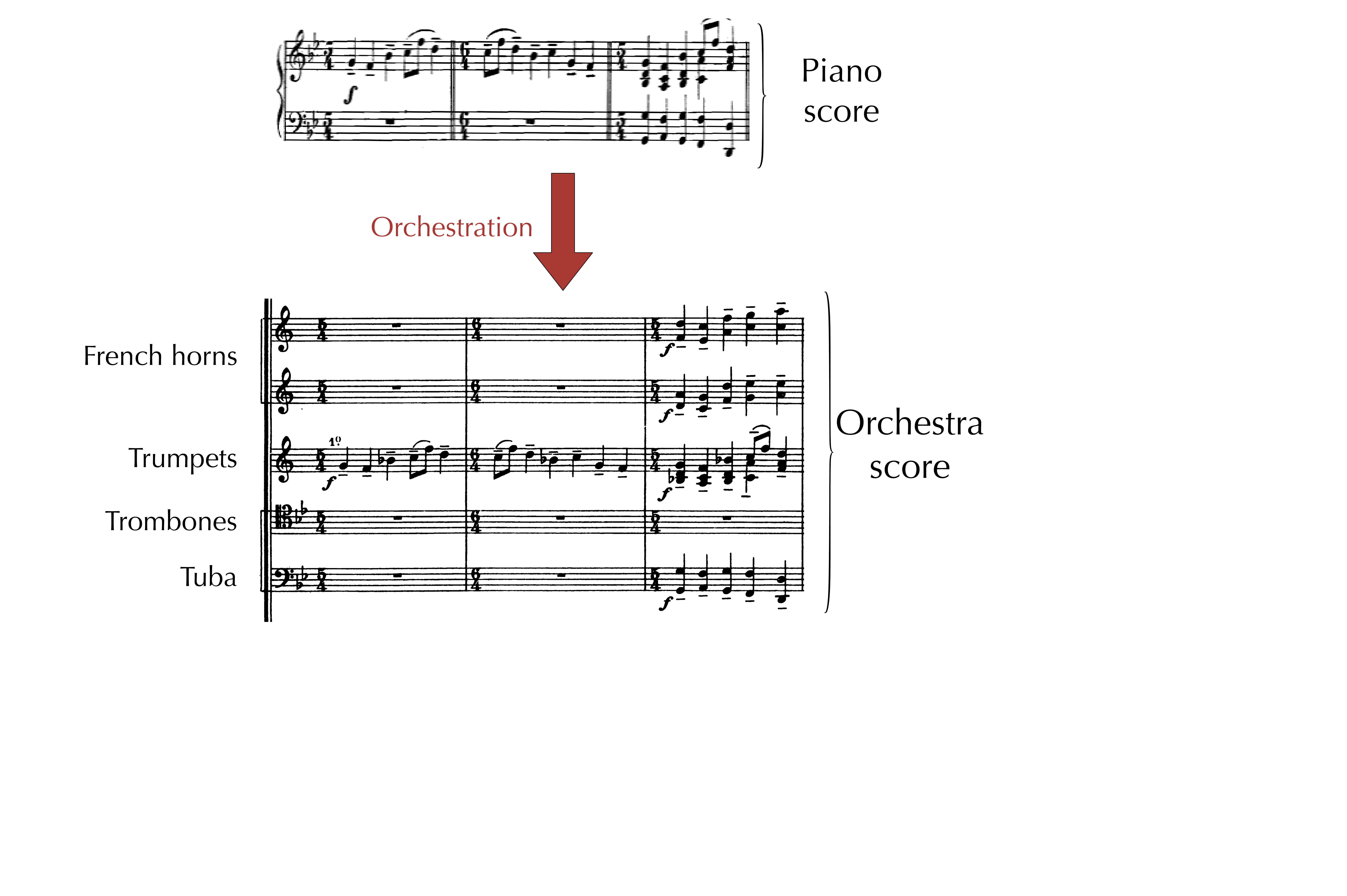}
\caption{\textit{Projective orchestration}. A piano score is projected on an orchestra. Even though a wide range of orchestrations exist for a given piano score, all of them will share strong relations with the original piano score. One given orchestration implicitly embeds the knowledge of the composer about timbre and orchestration.}
\label{fig:orch}
\end{figure}

Among the different orchestral writing techniques, one of them consists in first laying an harmonic and rhythmic structure in a piano score and then adding the orchestral timbre by spreading the different voices over the various instruments \cite{piston-orch}. We refer to this operation of extending a piano draft to an orchestral score as \textit{projective orchestration} \cite{eslingthesis}.
The orchestral repertoire contains a large number of such projective orchestrations (the piano reductions of Beethoven symphonies by Liszt or the \textit{Pictures at an exhibition}, a piano piece by Moussorgsky orchestrated by Ravel and other well-known composers). By observing an example of projective orchestration (Figure \ref{fig:orch}), we can see that this process involves more than the mere allocation of notes from the piano score across the different instruments. It rather implies harmonic enhancements and timbre manipulations to underline the already existing harmonic and rhythmic structure \cite{mcadams2013timbre}. However, the visible correlations between a piano score and its orchestrations appear as a fertile framework for laying the foundations of a computational exploration of orchestration.

Statistical inference offers a framework aimed at automatically extracting a structure from observations. These approaches hypothesize that a particular type of data is structured by an underlying probability distribution. The objective is to learn the properties of this distribution, by observing a set of those data. 
If the structure of the data is efficiently extracted and organized, it becomes then possible to generate novel examples following the learned distribution. A wide range of statistical inference models have been devised, among which \emph{deep learning} appears as a promising field \cite{bengio2013representation,LeCun:2015aa}. Deep learning techniques have been successfully applied to several musical applications and neural networks are now the state of the art in most music information retrieval \cite{humphrey2012moving,lee2011unsupervised,boulanger2013audio} and speech recognition \cite{hinton2012deep,DBLP:journals/corr/OordDZSVGKSK16} tasks. Several generative systems working with symbolic information (musical scores) have also been successfully applied to automatic music composition \cite{eck2002finding,lavrenko2003polyphonic,bosley2010learning,boulanger2012modeling,Johnson2015}
and automatic harmonization \cite{Sun}.

Automatic orchestration can actually cover a wide range of different applications.
In \cite{carpentier2010predicting,esling2010dynamic}, the objective is  to find the optimal combination of orchestral sounds in order to recreate any sonic target. An input of the system is a sound target, and the algorithm explores the different combination using a database of recorded acoustic instruments.
The work presented in \cite{pachet2016joyful} consists in modifying the style of an existing score. For instance, it can generate a \textit{bossa nova} version of a Beethoven's symphony.
To our best knowledge, the automatic projective orchestration task has only been investigated in \cite{handelman2012automatic} using a rule-based approach to perform an analysis of the piano score. Note that the analysis is automatically done, but not the allocation of the extracted structures to the different instruments.
Our approach is based on the hypothesis that the statistical regularities existing between a corpus of piano scores and their corresponding orchestrations could be uncovered through statistical inference. Hence, in our context, the data is defined as the scores, formed by a series of pitches and intensities for each instrument. The observations is a set of projective orchestrations performed by famous composers, and the probability distribution would model the set of notes played by each instrument conditionally on the corresponding piano score.

It might be surprising at first to rely solely on the symbolic information (scores) whereas orchestration is mostly defined by the spectral properties of instruments, typically not represented in the musical notation but rather conveyed in the signal information (audio recording).
However, we make the assumption that the orchestral projection performed by well-known composers effectively took into account the subtleties of timbre effects. Hence, spectrally consistent orchestrations could be generated by uncovering the composers' knowledge about timbre embedded in these scores.

Thus, we introduce the \emph{projective orchestration} task that is aimed at learning models able to generate orchestrations from unseen piano scores. We investigate a class of models called \textit{Conditional RBM} (cRBM) \cite{taylor2006modeling}. Conditional models implement a dependency mechanism that seems adapted to model the influence of the piano score over the orchestral score. In order to rank the different models, we establish a novel objective and quantitative evaluation framework, which is is a major difficulty for creative and systems. In the polyphonic music generation field, a predictive task with frame-level accuracy is commonly used by most systems \cite{boulanger2012modeling,lavrenko2003polyphonic,DBLP:journals/corr/YaoCVDD15}. However, we show that this frame-level accuracy is highly biased and maximized by models that simply repeat their last input. Hence, we introduce a novel event-level evaluation framework and benchmark the proposed models for projective orchestration. Then, we discuss the qualitative aspects of both the models and evaluation framework to explain the results obtained. Finally, we selected the most efficient model and implemented it in a system called \textit{Live Orchestral Piano} (LOP). This system performs real-time projective orchestration, allowing for anyone to play with an orchestra in real-time by simply playing on a MIDI keyboard.

The remainder of this paper is organized as follows. The first section introduces the state of the art in conditional models, in which RBM, cRBM and \textit{Factored-Gated cRBM} (\textit{FGcRBM}) models are detailed. Then, the projective orchestration task is presented along with an evaluation framework based on a event-level accuracy measure. The models are evaluated within this framework and compared to existing models. Then, we introduce \textit{LOP}, the real-time projective orchestration system. Finally, we provide our conclusions and directions of future work.

\section{Conditional neural networks}
In this section, three statistical inference models are detailed. The \textit{RBM}, \textit{cRBM} and \textit{FGcRBM} are presented by increasing level of complexity.
\subsection{Restricted Boltzmann Machine}
\subsubsection{An energy based model}
The Restricted Boltzmann Machine (\textit{RBM}) is a graphical probabilistic model defined by stochastic visible units $\bm{v} = \{ v_1, .., v_{n_v} \}$ and hidden units $\bm{h} = \{ h_1, .., h_{n_h} \}$. The visible and hidden units are tied together by a set of weights $\bm{W}$ following the conditional probabilities
\begin{align}
\label{eq:conditional_rbm}
p(v_i=1|\bm{h}) = \sigma(a_i + \sum_{j=1}^{n_h} W_{ij} h_j)\\
p(h_j=1|\bm{v}) = \sigma(b_j + \sum_{i=1}^{n_v} W_{ij} v_i)
\end{align}
where $\sigma(x) = \frac{1}{1 + exp(-x)}$ is the sigmoid function.

The joint probability of the visible and hidden variables is given by
\begin{equation}
p_{model}(\bm{v},\bm{h}) =  \frac{\exp^{-E(\bm{v},\bm{h})}}{Z}
\end{equation}
where
\begin{equation}
\label{eq:energy}
E(\bm{v},\bm{h}) = - \sum_{i=1}^{n_{v}} a_i v_{i}  - \sum_{i=1}^{n_v} \sum_{j=1}^{n_h} v_{i} W_{ij} h_{j} - \sum_{j = 1}^{n_h} b_j h_{j}
\end{equation}
with $\bm{\Theta} = \left\{\bm{W},\bm{b},\bm{a}\right\}$ the parameters of the network.

\subsubsection{Training procedure}
The values of the parameters $\bm{\Theta}$ of a \textit{RBM} are learned through the minimization of an error function, usually defined as the negative log-likelihood
\begin{equation}
\label{eq:likelihood}
\mathcal{L(\bm{\theta}|\mathcal{D})}  = \frac{1}{N_{\mathcal{D}}} \sum_{\bm{v^{(l)}} \in \mathcal{D}} - \ln \left( p(\bm{v^{(l)}}|\bm{\theta})\right)
\end{equation}
where $\mathcal{D}$ is the training dataset and $N_{\mathcal{D}}$ the number of elements that it contains.

The training procedure then modifies the parameters $\Theta$ of the model by using the gradient of the error function. However, the gradient of the negative log-likelihood is intractable.
Therefore, an approximation of this quantity is obtained by running a Gibbs sampling chain, a procedure known as the Contrastive Divergence (\textit{CD}) algorithm \cite{Fischer2012,hinton2010practical}. The method alternates between sampling the conditional probabilities of the visible units while keeping the hidden units fixed (Equation \ref{eq:conditional_rbm}) and then doing the opposite, until the vector of visible units is close to the real distribution of the model. \textit{CD} provides an adequate approximation for minimizing the loss function and guarantees that the Kullback-Leibler divergence between the model and data distribution is reduced after each iteration\cite{hinton2002training}.

\subsection{Conditional models}
Conditional models introduce \emph{context units}, which provides an interesting way of influencing the learning. We briefly present the \textit{cRBM} and \textit{FGcRBM} models and redirect interested readers to the original paper \cite{taylor2009factored} for more details.

\subsubsection{Conditional RBM}
In the \textit{cRBM} model, the influence of the context units is implemented by introducing an additive term on the biases $a_i$ and $b_j$ of both visible and hidden units.
\begin{align}
\tilde{a}_i = a_i + \sum_{k=1}^{n_x} A_{ki} x_k\\
\tilde{b}_j = b_j + \sum_{k=1}^{n_x} B_{kj} x_k
\end{align}
where $\tilde{\bm{a}}$ and $\tilde{\bm{b}}$ are called \emph{dynamic biases} (by opposition to the static biases $\bm{a}$ and $\bm{b}$). The additive term is a linear combination of a third set of random variables called context units $\bm{x} = \{ x_1, .., x_k, .., x_{n_x} \}$.
By learning the matrices of parameters $\bm{A}$ and $\bm{B}$, the context units can stimulate or inhibit the hidden or visible units depending on the input. Hence, this provides a mechanism that can influence the generation based on a particular context. After replacing the static biases by dynamic ones, the conditional distributions (Equation \ref{eq:conditional_rbm}), energy function (Equation \ref{eq:energy}) and training procedure remains the same as for the \textit{RBM}.

\subsubsection{Factored Gated cRBM}
The Factored Gated cRBM (\textit{FGcRBM}) model  \cite{taylor2009factored} proposes to extend the cRBM model by adding a layer of feature units $\bm{z}$, which modulates the weights of the conditional architecture in a multiplicative way. The dynamic biases of the visible and hidden units are defined by
\begin{align}
\hat{a}_{i} = a_{i} + \sum_{f} \sum_{kl}A_{if}A_{kf}A_{lf}x_{k}z_{l}\\
\hat{b}_{j} = b_{j} + \sum_{f} \sum_{kl}B_{jf}B_{kf}B_{lf}x_{k}z_{l}
\end{align}
and the energy function by
\begin{align}
E(\bm{v},\bm{h}) &= - \sum_i \tilde{a}_i v_i \\ 
& - \sum_f \sum_{ijl}  W_{if} W_{jf} W_{lf} v_i h_j z_l - \sum_j \tilde{b}_j h_j
\end{align}

This multiplicative influence can be interpreted as a modification of the energy function of the model depending on the latent variables $\bm{z}$. For a fixed configuration of feature units, a new energy function is defined by the \textit{cRBM} ($\bm{v}$, $\bm{h}$, and $\bm{x}$).
Ideally, the three way interactions could be modeled by three dimensional tensors, such as  $W_{ijl}$. To define this, the number of parameters should grow cubically with the number of units. Hence, to reduce the computation load, the three-dimensional tensors are factorized into a product of three matrices by including factor units indexed by $f$ such that $W_{ijl} = W_{if} . W_{jf} . W_{lf}$.

\section{Projective orchestration}
In this section, we introduce and formalize the automatic projective orchestration task presented in Figure \ref{fig:orch}. Then, the piano-roll representation used to process the scores is detailed. Finally, the application of the previously introduced models in this particular context is defined.

\subsection{Task formalization}
In the general setting, a piano score and an orchestral score can be represented as a sequence of states $P(t)$ and $O(t)$. We consider that the time instants of both sequences are aligned. The projective orchestration task consists in predicting the present orchestral state given the present piano state and the past orchestral states
\begin{equation}
\hat{O}(t) = f\left(P(t), O(t-N),... ,O(t-1)\right)
\end{equation}
where $N$ is a parameter representing the temporal order (or \textit{horizon}) of the task.

\subsection{Data representation}
\begin{figure}[ht]
\centering
\includegraphics[scale=0.37]{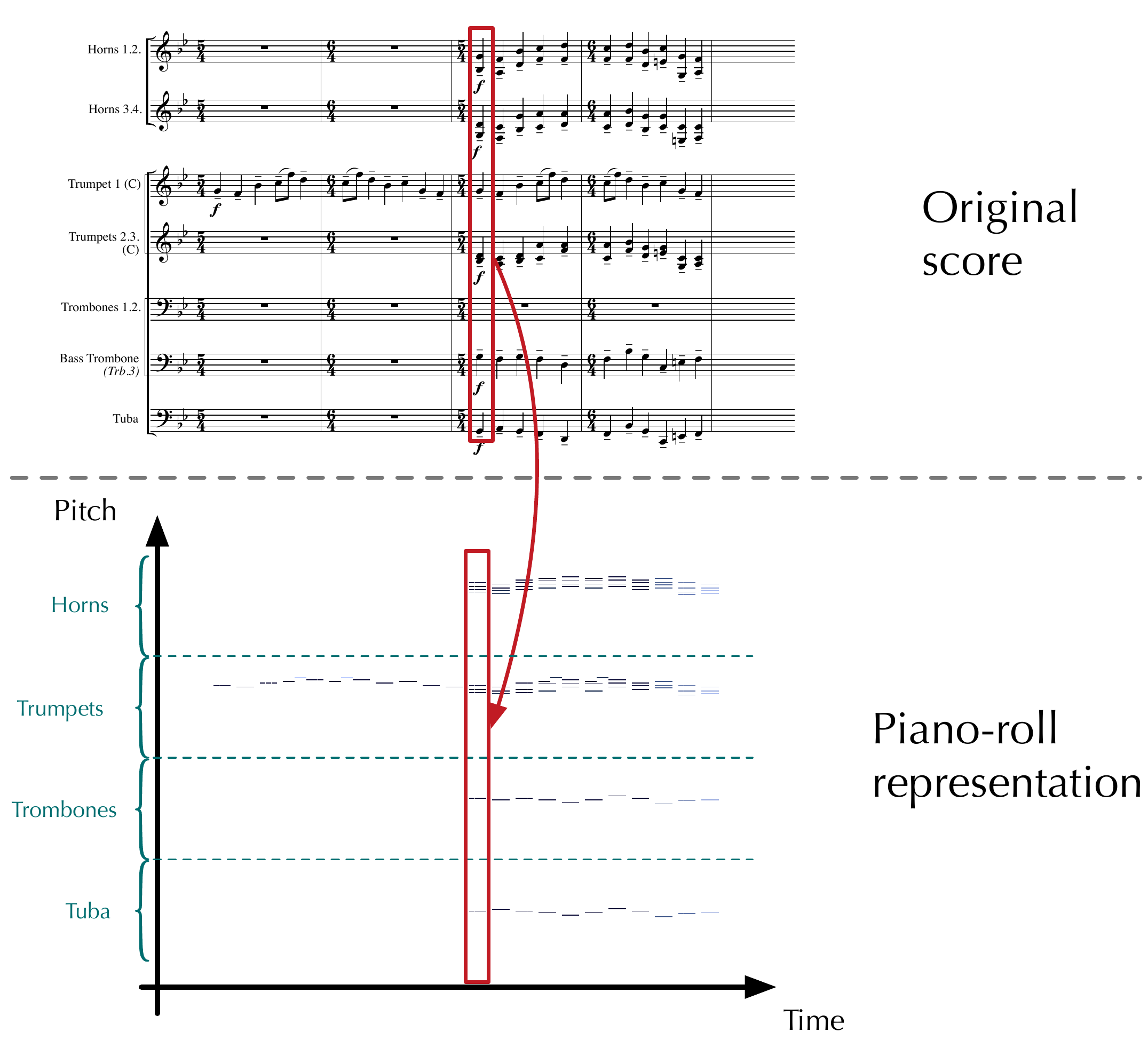}
\caption{From the score of an orchestral piece, a convenient \textit{piano-roll} representation is extracted. A piano-roll $pr$ is a matrix whose rows represent pitches and columns represent a time frame depending on the time quantization. A pitch $p$ at time $t$ played with an intensity $i$ is represented by $pr(p,t) = i$, $0$ being a note off. This definition is extended to an orchestra by simply concatenating the \textit{piano-rolls} of every instruments along the pitch dimension.}
\label{fig:piano-roll}
\end{figure}

A \textit{piano-roll} representation is used to process both the piano and orchestral scores. This representation is commonly used to model a single polyphonic instrument (Figure~\ref{fig:piano-roll}).
Its extension to orchestral scores is obtained straightforwardly by concatenating the \textit{piano-rolls} of each instrument along the pitch dimension in order to obtain a matrix. The rhythmic quantization is defined as the number of time frame in the \textit{piano-roll} per quarter note. Hence, we define the sequence of piano and orchestra states $P(t)$ and $O(t)$ as the sequence of the column vectors formed by the \textit{piano-roll} representations, where $t$ is a discrete time index.

\subsection{Model definition}
The specific implementation of the models for the projective orchestration task is detailed in this subsection.
\begin{figure*}
 	\begin{centering}
 		\includegraphics[scale=0.2]{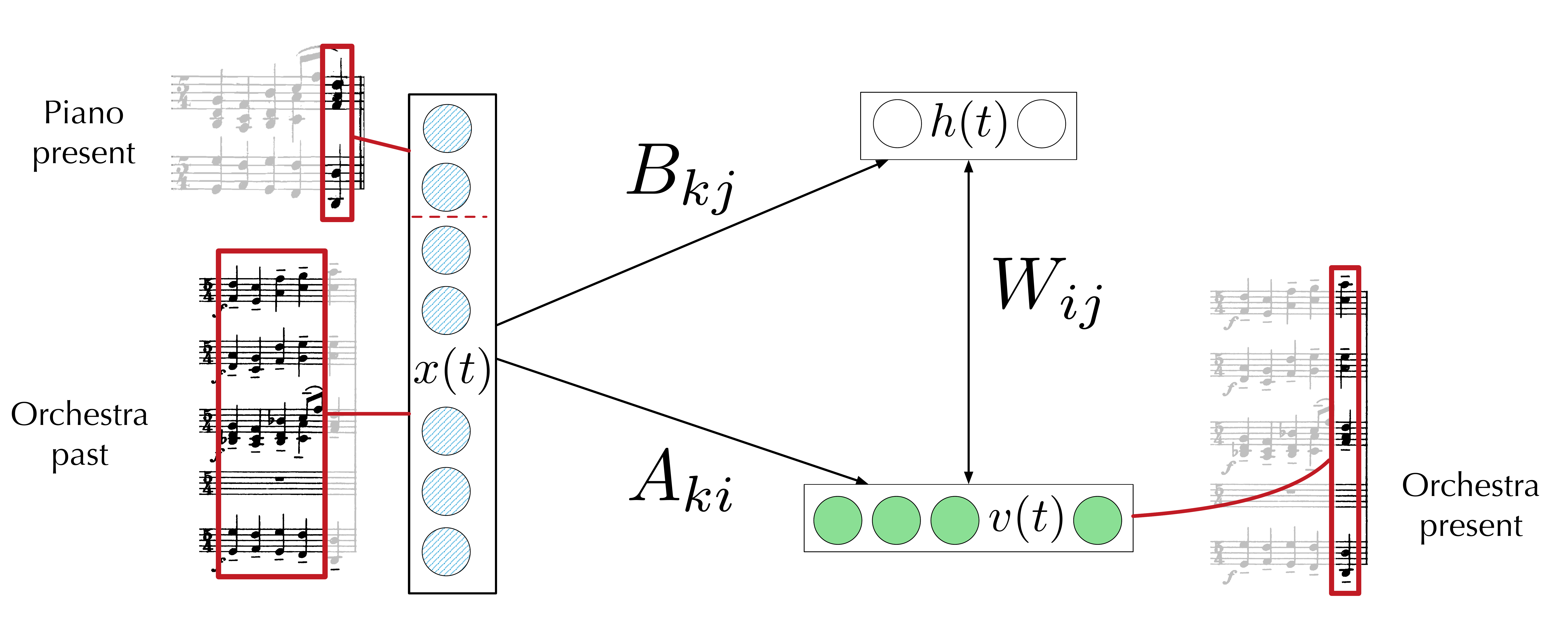}
 		\par\end{centering}
 	\caption{Conditional Restricted Boltzmann Machine in the context of projective orchestration. The visible units $\bm{v}$ represent the present orchestral frame $O(t)$. The context units $\bm{x}$ represents the concatenation of the past orchestral and present piano frames $[P(t), O(t-N), ..., O(t-1)]$. 
After training the model, the generation phase consists in clamping the context units (blue hatches) to their known values and sampling the visible units (filled in green) by running a Gibbs sampling chain.}
 	\label{fig:lop_models}
 \end{figure*}
\subsubsection{RBM}
In the case of projective orchestration, the visible units are defined as the concatenation of the past and present orchestral states and the present piano state
\begin{equation}
\label{eq:visible_rbm}
\bm{v} = \left[P(t),O(t-N),...,O(t)\right]
\end{equation}

Generating values from a trained \textit{RBM} consists in sampling from the distribution $p(\bm{v})$ it defines. Since it is an intractable quantity, a common way of generating from a \textit{RBM} is to perform $K$ steps of Gibbs sampling, and consider that the visible units obtained at the end of this process will represent an adequate approximation of the true distribution $p(\bm{v})$.
The quality of this approximation increases with the number $K$ of Gibbs sampling steps.

In the case of projective orchestration, the vectors $P(t)$ and $[O(t-N), ..., O(t-1)]$ are known, and only the vector $O(t)$ has to be generated. 
Thus, it is possible to infer the approximation $\hat{O}(t)$ by clamping $O(t-N), ..., O(t-1)$ and $P(t)$ to their known values and using the CD algorithm. This technique is known as \textit{inpainting} \cite{Fischer2012}.

However, extensive efforts are made during the training phase to infer values that are finally clamped during the generation phase. Conditional models offer a way to separate the context (past orchestra and present piano) from the data we actually want to generate (present orchestra).

\subsubsection{cRBM and FGcRBM}
In our context, we define the \textit{cRBM} units as
\begin{align*}
\bm{v} &= O(t) \\
\bm{x} &= \left[P(t), O(t-N), ..., O(t-1)\right]
\end{align*}
and the \textit{FGcRBM} units as
 \begin{align*}
 \bm{v} &= O(t) \\
 \bm{x} &= \left[O(t-N), ..., O(t-1)\right]\\
 \bm{z} &= P(t)
 \end{align*}
 
Generating the orchestral state $O(t)$ can be done by sampling visible units from those two models. This is done by performing $K$ Gibbs sampling steps, while clamping the context units (piano present and orchestral past) in the case of \textit{cRBM} (as displayed in Figure~\ref{fig:lop_models}) and both the context (orchestral past) and latent units (piano present) in the case of the \textit{FGcRBM} to their known values.

\subsubsection{Dynamics}
As aforementioned, we use binary stochastic units, which solely indicate if a note is on or off. Therefore, the velocity information is discarded. Although we believe that the velocity information is of paramount importance in orchestral works (as it impacts the number and type of instruments played), this first investigation provides a valuable insight to determine the architectures and mechanisms that are the most adapted to this task.

\subsubsection{Initializing the orchestral past}
Gibbs sampling requires to initialize the value of the visible units.
During the training process, they can be set to the known visible units to reconstruct, which speeds up the convergence of the Gibbs chain.
For the generation step, the first option we considered was to initialize the visible units with the previous orchestral frame $O(t-1)$. However, because repeated notes are very common in the training corpus, the negative sample obtained at the end of the Gibbs chain was often the initial state itself $\hat{O}(t) = O(t-1)$.
Thus, we initialize the visible units by sampling a uniform distribution between 0 and 1.

\section{Evaluation framework}
In order to assess the performances of the different models presented in the previous section, we introduce a quantitative evaluation framework for the projective orchestration task. Hence, this first requires to define an accuracy measure to compare a predicted state $\hat{O}(t)$ with the ground-truth state $O(t)$ written in the orchestral score. As we experimented with accuracy measures commonly used in the music generation field \cite{DBLP:journals/corr/YaoCVDD15,boulanger2012modeling,lavrenko2003polyphonic}, we discovered that these are heavily biased towards models that simply repeat their last input. Hence, we discuss the temporal granularities used to process the scores as piano-rolls and propose two alternatives that we call \textit{frame-level} and \textit{event-level} accuracy measures.

\subsection{Accuracy measure}
In order to discriminate the performances of different generative models, the ideal evaluation would be to compute the likelihood of an unseen set of data. However, we have seen that this quantity is intractable for probabilistic models such as the \textit{RBM}, \textit{cRBM} and \textit{FGcRBM}. An alternative criterion commonly used in the music generation field \cite{DBLP:journals/corr/YaoCVDD15,boulanger2012modeling,lavrenko2003polyphonic} is the accuracy measure defined as 
\begin{equation}
\text{Accuracy}  = \frac{TP(t)}{TP(t) + FP(t) + FN(t)}
\label{eq:accuracy}
\end{equation}
where the true positives $TP(t)$ is the number of notes correctly predicted, the false positives $FP(t)$ is the number of notes predicted which are not in the original sequence and the false negatives $FN(t)$  is the number on unreported notes.

\subsubsection{Temporal granularities}
 \begin{figure}
\centering
\includegraphics[scale=0.42]{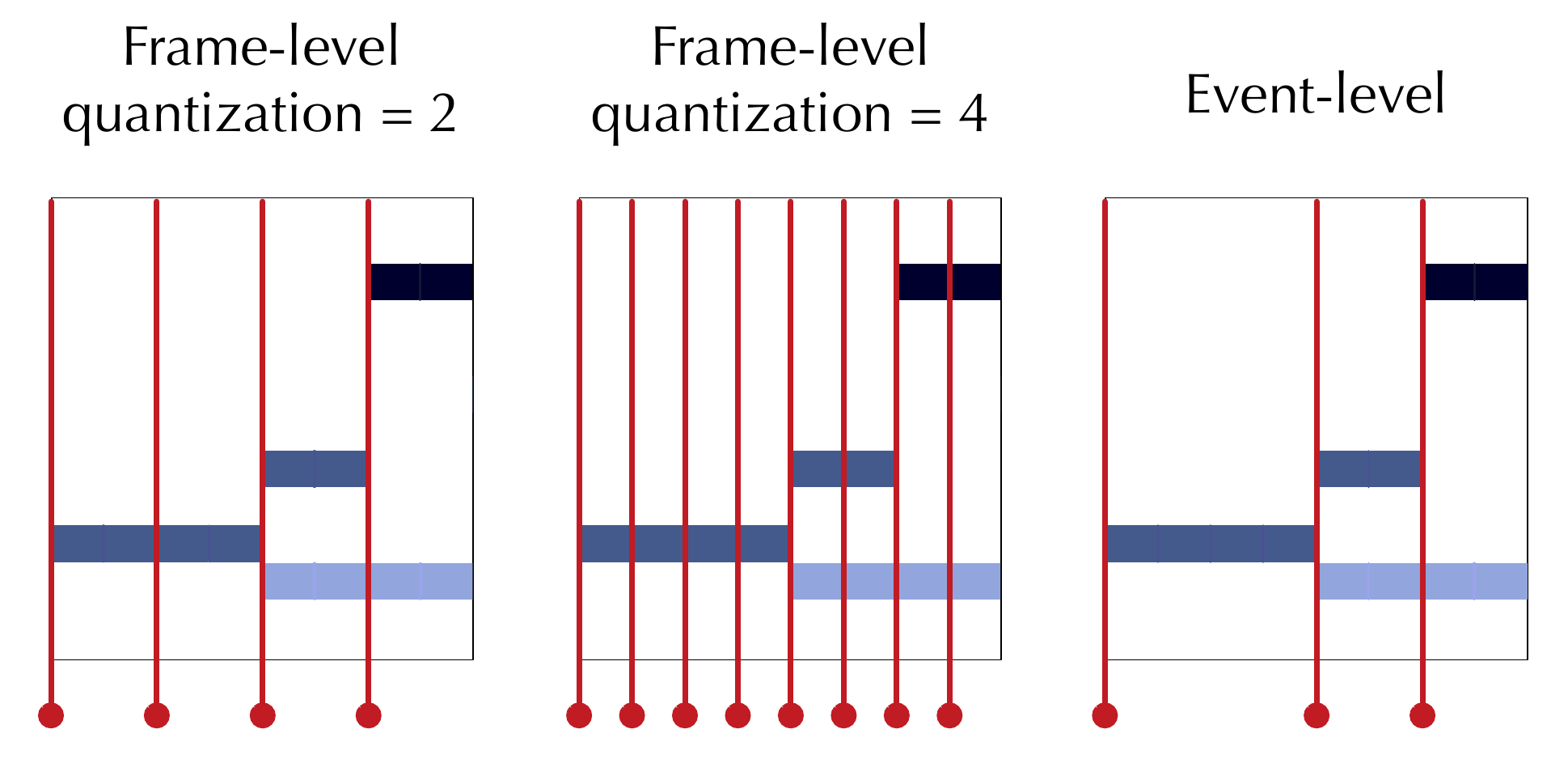}
\caption{\textit{Frame-level} and \textit{event-level} granularities. In the case of the frame-level granularity (left and middle), we can see that, as the rhythmic quantization gets finer, an increasing number of consecutive frames become identical. However, with the event-level granularity, the bias imposed by the temporal quantization can be alleviated, while still accounting for events that are truly repeated in the score.}
\label{fig:frame_vs_event}
\end{figure}
When the accuracy measure defined in Equation~\ref{eq:accuracy} is computed for each time index $t$ (termed here \textit{frame-level} granularity) of the \textit{piano-rolls}, this accuracy highly depends on the rhythmic quantization.
Indeed, it can be observed on the figure \ref{fig:frame_vs_event} that when the quantization used to compute the \textit{piano-roll} gets finer, an increasing number of successive states in the sequence become identical.
As a consequence, a model which simply predicts the state $\hat{O}(t)$ by repeating the previous state $O(t-1)$ gradually becomes the best model as the quantization gets finer.

To alleviate this problem, we rely on an \textit{event-level} granularity, which only assesses the prediction for frames of the piano-roll where an event occurs. We define an event as a time $t_{e}$ where $\text{Orch}(t_{e}) \neq \text{Orch}(t_{e} - 1)$.
Hence, the temporal precision of the event-level representation no longer depends on a quantization parameter and the scores are seen as a succession of events with no rhythmic structure. In the general case, this measure should be augmented with the notion of the event durations. However, in our case, the rhythmic structure of the projected orchestral score is imposed by the original piano score.

In the following section, we discuss the impact of the frame-level and event-level measures on the performance of the different models.

\section{Results}
\label{sec:results}
\subsection{Database}
We created a specific MIDI database of piano scores and their corresponding orchestrations performed by famous composers. A total of 223 pairs of piano scores and corresponding orchestrations have been collected and 28 different instruments are represented.
The database is freely available\footnote{\url{https://qsdfo.github.io/LOP/database}}, along with detailed statistics and the source code to reproduce our results.

Following standard notations, all instruments of the same section are grouped under the same part in the score. For instance, the \textit{violins}, which might be played by several instrumentalists, is written as a single part.
Besides, in order to reduce the number of units, we systematically remove, for each instrument, any pitch which is never played in the database. Hence, the dimension of the orchestral vector is reduced from 3584 to 1220.

The dataset is split between train, validation and test sets of files that represent respectively 80\%, 10\% and 10\% of the full set.
For reproducibility, the different sets we used are provided as text files on the companion website.

\subsection{Quantitative evaluation}
We evaluate five different models on the projective orchestration task. The first model is a random generation of the orchestral frames from a Bernoulli distribution of parameter $0.5$ in order to evaluate the complexity of the task. The second model predicts an orchestral frame at time $t$ by repeating the frame at time $t-1$, in order to evaluate the \textit{frame-level} bias. These two naive models constitute a baseline against which we compare the \textit{RBM}, \textit{cRBM} and \textit{FGcRBM} models.

The complete implementation details and hyper-parameters can be found on the companion website.

\begin{table}[h]
	\centering
    \scalebox{0.85}{
	\begin{tabular}{c c c c}
		\hline
		\thead{Model} & \thead{Frame-level\\ accuracy (Q = 4)} & \thead{Frame-level\\ accuracy (Q = 8)} & \thead{Event-level\\ accuracy} \\
		\hline
		Random & 0.73 & 0.73 & 0.72\\ 
		Repeat & 61.79 & 76.41 & 50.70\\
		\hline \hline
		RBM & 7.67 & 4.56 & 1.39\\ 
		cRBM & 5.12 & 34.25 & 27.67\\ 
		FGcRBM & 33.86 & 43.52 & 25.80\\
	\end{tabular}
    }
	\caption{Results of the different models for the projective orchestration task based on frame-level accuracies with a quantization of 4 and 8 and event-level accuracies.}
	\label{tab:result_event_level}
\end{table}

The results are summarized in Table~\ref{tab:result_event_level}. In the case of frame-level accuracies, the \textit{FGcRBM} provides the highest accuracy amongst probabilistic models. However, the repeat model remains superior to all models in that case. If we increase the quantization, we see that the performances of the repeat model increase considerably. This is also the case for the \textit{cRBM} and \textit{FGcRBM} models as the predictive objective becomes simpler.

The RBM model obtains poor performances for all the temporal granularities and seems unable to grasp the underlying structure.
Hence, it appears that the conditioning is of paramount importance to this task. This could be explained by the fact that a dynamic temporal context is one of the foremost properties in musical orchestration.

The \textit{FGcRBM} has slightly worse performances than the \textit{cRBM} in the event-level framework. 
The introduction of three ways interactions might not be useful or too intricate in the case of projective orchestration. Furthermore, disentangling the interactions of the piano and the orchestra against the influence of the temporal context might not be clearly performed by the factored model because of the limited size of the database. Nevertheless, it seems that conditional probabilistic models are valid candidates to tackle the projective orchestration task.

Even with the event-level accuracy, the repeat model still obtains a very strong score, which further confirms the need to devise more subtle accuracy measures. This result directly stems from the properties of orchestral scores, where it is common that several instruments play a sustained background chord, while a single solo instrument performs a melody. Hence, even in the event-level framework, most of the notes will be repeated between two successive events. This can be observed on the database section of the companion website.

\subsection{Qualitative analysis}
In order to allow for a qualitative analysis of the results, we provide several generated orchestrations from different models on our companion website. We also report the different accuracy scores obtained by varying the hyper-parameters. In this analysis, it appeared that the number of hidden units and number of Gibbs sampling steps are the two crucial quantities. For both the \textit{cRBM} and \textit{FGcRBM} models, best results are obtained when the number of hidden units is over 3000 and the number of Gibbs sampling steps larger than 20.

We observed that the generated orchestrations from the \textit{FGcRBM} might sound better than the ones generated by the \textit{cRBM}.
Indeed, it seems that the \textit{cRBM} model is not able to constraint the possible set of notes to the piano input. To confirm this hypothesis, we computed the accuracy scores of the models while setting the whole piano input to zero. By doing this, we can evaluate how much a model rely on this information. The score for those \textit{corrupted} inputs considerably dropped for the \textit{FGcRBM} model, from 25.80\% to 1.15\%. This indicates that the prediction of this model heavily rely on the piano input. Instead, the corrupted score of the \textit{cRBM} drops from 27.67\% to 17.80\%. This shows that the piano input is not considered as a crucial information for this model, which rather performs a purely predictive task based on the orchestral information.

\subsection{Discussion}
It is important to state that we do not consider this prediction task as a reliable measure of the creative performance of different models. Indeed, predicting and creating are two fairly different things. Hence, the predictive evaluation framework we have built does not assess the generative ability of a model, but is rather used as a selection criterion among different models. Furthermore, this provides an interesting \textit{auxiliary task} towards orchestral generation.

\section{Live Orchestral Piano (LOP)}
We introduce here the \emph{Live Orchestral Piano} (LOP), which allows to compose music with a full classical orchestra in real-time by simply playing on a MIDI piano. This framework relies on the knowledge learned by the models introduced in the previous sections in order to perform the projection from a piano melody to the orchestra.

\subsection{Workflow}
\begin{figure*}
\begin{centering}
\includegraphics[scale=0.55]{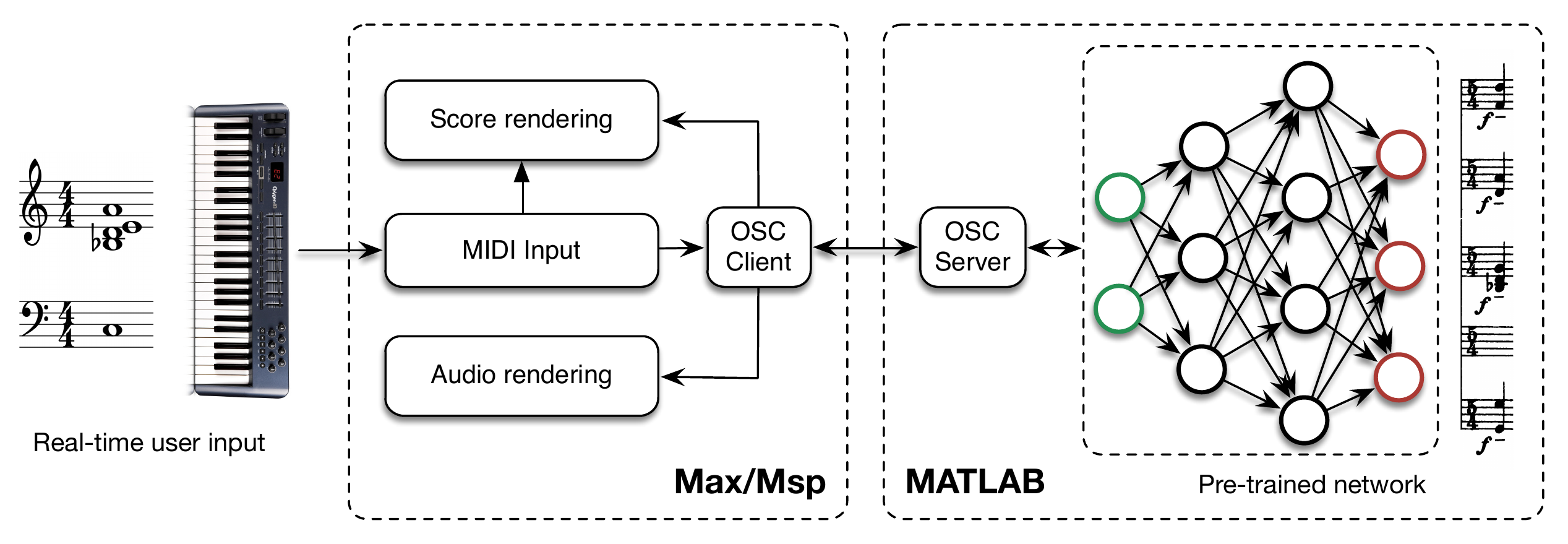}
\par\end{centering}
\caption{\label{fig:Live-orchestral-piano}Live orchestral piano (LOP) implementation workflow. The user plays on a MIDI piano which is transcribed and sent via OSC from the Max/Msp client. The MATLAB server
uses this vector of notes and process it following the aforementioned
models in order to obtain the corresponding orchestration. This information
is then sent back to Max/Msp which performs the real-time audio rendering. }
\end{figure*}

The software is implemented on a client/server paradigm. This choice
allows to separate the orchestral projection task from the interface
and sound rendering engine. That way, multiple interfaces can easily
be implemented. Also, it should be noted that separating the computing
and rendering aspects on different computers could allow to use high-quality
and CPU-intensive orchestral rendering plugins, while ensuring the real-time constraint on the overall system (preventing degradation on the computing part). The complete workflow is presented in Figure~\ref{fig:Live-orchestral-piano}.

The user plays a melody (single notes or chords sequences) with a MIDI keyboard, which is retrieved inside the interface. The interface has been developed in Max/Msp, to facilitate both the score and audio rendering aspects. The interface transmits this symbolic information (as a variable-length vector of active notes) via OSC to the MATLAB server. This interface also performs a transcription of the piano score to the screen, by relying on the \emph{Bach} library environment. The server uses this vector of events to produce an 88 vector of binary note activations. This vector is then processed by the orchestration models presented in the previous sections in order to obtain a projection of a specific symbolic piano melody to the full orchestra. The resulting orchestration is then sent back to the client interface which performs both the real-time audio rendering and score transcription. The interface also provides a way to easily switch between different models, while controlling other hyper-parameters of the sampling

\section{Conclusion and future works}
We have introduced a system for real-time orchestration of a midi piano input. First, we formalized the projective orchestration task and proposed an evaluation framework that could fit the constraints of our problem. We showed that the commonly used \textit{frame-level} accuracy measure is highly biased towards repeating models and proposed an \textit{event-level} measure instead. Finally, we assessed the performance of different probabilistic models and discussed the better performances of the FGcRBM model.

The conditional models have proven to be effective in the orchestral inference evaluation framework we defined. The observations of the orchestration generated by the models tend to confirm these performance scores, and support the overall soundness of the framework. However, we believe that the performances can be further improved by mixing conditional and recurrent models.

The most crucial improvement to our system would be to include the dynamics of the notes. Indeed, many orchestral effects are directly correlated to the intensity variations in the original piano scores.  Hence, by using binaries units, an essential information is discarded.
Furthermore, the sparse representation of the data suggests that a more compact distributed representation might be found. Lowering the dimensionality of the data would greatly improve the efficiency of the learning procedure. For instance, methods close to the word-embedding techniques used in natural language processing might be useful \cite{kiros2015skip}.

The general objective of building a generative model for time series is one of the currently most prominent topic for the machine learning field. Orchestral inference sets a slightly more complex framework where a generated multivariate time series is conditioned by an observed time series (the piano score). Finally, being able to grasp long-term dependencies structuring orchestral pieces appears as a promising task.



\section{Acknowledgements}
This work has been supported by the \textit{NVIDIA GPU Grant} program.
The authors would like to thank Denys Bouliane for kindly providing a part of the orchestral database.


\bibliography{biblio}

\begin{thebibliography}{10}
\providecommand{\url}[1]{#1}
\csname url@samestyle\endcsname
\providecommand{\newblock}{\relax}
\providecommand{\bibinfo}[2]{#2}
\providecommand{\BIBentrySTDinterwordspacing}{\spaceskip=0pt\relax}
\providecommand{\BIBentryALTinterwordstretchfactor}{4}
\providecommand{\BIBentryALTinterwordspacing}{\spaceskip=\fontdimen2\font plus
\BIBentryALTinterwordstretchfactor\fontdimen3\font minus
  \fontdimen4\font\relax}
\providecommand{\BIBforeignlanguage}[2]{{%
\expandafter\ifx\csname l@#1\endcsname\relax
\typeout{** WARNING: IEEEtran.bst: No hyphenation pattern has been}%
\typeout{** loaded for the language `#1'. Using the pattern for}%
\typeout{** the default language instead.}%
\else
\language=\csname l@#1\endcsname
\fi
#2}}
\providecommand{\BIBdecl}{\relax}
\BIBdecl

\bibitem{koechli_orch}
C.~Koechlin, \emph{Trait{\'e} de l'orchestration}.\hskip 1em plus 0.5em minus
  0.4em\relax {\'E}ditions Max Eschig, 1941.

\bibitem{Rimsky-Korsakov:1873aa}
N.~Rimsky-Korsakov, \emph{Principles of Orchestration}.\hskip 1em plus 0.5em
  minus 0.4em\relax Russischer Musikverlag, 1873.

\bibitem{mcadams2013timbre}
S.~McAdams, ``Timbre as a structuring force in music,'' in \emph{Proceedings of
  Meetings on Acoustics}, vol.~19, no.~1.\hskip 1em plus 0.5em minus
  0.4em\relax Acoustical Society of America, 2013, p. 035050.

\bibitem{tardieu2012perception}
D.~Tardieu and S.~McAdams, ``Perception of dyads of impulsive and sustained
  instrument sounds,'' \emph{Music Perception}, vol.~30, no.~2, pp. 117--128,
  2012.

\bibitem{piston-orch}
W.~Piston, \emph{Orchestration}.\hskip 1em plus 0.5em minus 0.4em\relax New
  York: Norton, 1955.

\bibitem{eslingthesis}
P.~Esling, ``Multiobjective time series matching and classification,'' Ph.D.
  dissertation, IRCAM, 2012.

\bibitem{bengio2013representation}
Y.~Bengio, A.~Courville, and P.~Vincent, ``Representation learning: A review
  and new perspectives,'' \emph{Pattern Analysis and Machine Intelligence, IEEE
  Transactions on}, vol.~35, no.~8, pp. 1798--1828, 2013.

\bibitem{LeCun:2015aa}
\BIBentryALTinterwordspacing
Y.~LeCun, Y.~Bengio, and G.~Hinton, ``Deep learning,'' \emph{Nature}, vol. 521,
  no. 7553, pp. 436--444, 05 2015. [Online]. Available:
  \url{http://dx.doi.org/10.1038/nature14539}
\BIBentrySTDinterwordspacing

\bibitem{humphrey2012moving}
E.~J. Humphrey, J.~P. Bello, and Y.~LeCun, ``Moving beyond feature design: Deep
  architectures and automatic feature learning in music informatics.'' in
  \emph{ISMIR}.\hskip 1em plus 0.5em minus 0.4em\relax Citeseer, 2012, pp.
  403--408.

\bibitem{lee2011unsupervised}
H.~Lee, R.~Grosse, R.~Ranganath, and A.~Y. Ng, ``Unsupervised learning of
  hierarchical representations with convolutional deep belief networks,''
  \emph{Communications of the ACM}, vol.~54, no.~10, pp. 95--103, 2011.

\bibitem{boulanger2013audio}
N.~Boulanger-Lewandowski, Y.~Bengio, and P.~Vincent, ``Audio chord recognition
  with recurrent neural networks.'' in \emph{ISMIR}.\hskip 1em plus 0.5em minus
  0.4em\relax Citeseer, 2013, pp. 335--340.

\bibitem{hinton2012deep}
G.~Hinton, L.~Deng, D.~Yu, G.~E. Dahl, A.-r. Mohamed, N.~Jaitly, A.~Senior,
  V.~Vanhoucke, P.~Nguyen, T.~N. Sainath \emph{et~al.}, ``Deep neural networks
  for acoustic modeling in speech recognition: The shared views of four
  research groups,'' \emph{IEEE Signal Processing Magazine}, vol.~29, no.~6,
  pp. 82--97, 2012.

\bibitem{DBLP:journals/corr/OordDZSVGKSK16}
\BIBentryALTinterwordspacing
A.~van~den Oord, S.~Dieleman, H.~Zen, K.~Simonyan, O.~Vinyals, A.~Graves,
  N.~Kalchbrenner, A.~W. Senior, and K.~Kavukcuoglu, ``Wavenet: {A} generative
  model for raw audio,'' \emph{CoRR}, vol. abs/1609.03499, 2016. [Online].
  Available: \url{http://arxiv.org/abs/1609.03499}
\BIBentrySTDinterwordspacing

\bibitem{eck2002finding}
D.~Eck and J.~Schmidhuber, ``Finding temporal structure in music: Blues
  improvisation with lstm recurrent networks,'' in \emph{Neural Networks for
  Signal Processing, 2002. Proceedings of the 2002 12th IEEE Workshop
  on}.\hskip 1em plus 0.5em minus 0.4em\relax IEEE, 2002, pp. 747--756.

\bibitem{lavrenko2003polyphonic}
V.~Lavrenko and J.~Pickens, ``Polyphonic music modeling with random fields,''
  in \emph{Proceedings of the eleventh ACM international conference on
  Multimedia}.\hskip 1em plus 0.5em minus 0.4em\relax ACM, 2003, pp. 120--129.

\bibitem{bosley2010learning}
S.~Bosley, P.~Swire, R.~M. Keller \emph{et~al.}, ``Learning to create jazz
  melodies using deep belief nets,'' 2010.

\bibitem{boulanger2012modeling}
N.~Boulanger-Lewandowski, Y.~Bengio, and P.~Vincent, ``Modeling temporal
  dependencies in high-dimensional sequences: Application to polyphonic music
  generation and transcription,'' \emph{arXiv preprint arXiv:1206.6392}, 2012.

\bibitem{Johnson2015}
D.~Johnson, ``hexahedria,''
  \url{http://www.hexahedria.com/2015/08/03/composing-music-with-recurrent-neural-networks/},
  08 2015.

\bibitem{Sun}
F.~Sun, ``Deephear - composing and harmonizing music with neural networks,''
  \url{http://web.mit.edu/felixsun/www/neural-music.html}.

\bibitem{carpentier2010predicting}
G.~Carpentier, D.~Tardieu, J.~Harvey, G.~Assayag, and E.~Saint-James,
  ``Predicting timbre features of instrument sound combinations: Application to
  automatic orchestration,'' \emph{Journal of New Music Research}, vol.~39,
  no.~1, pp. 47--61, 2010.

\bibitem{esling2010dynamic}
P.~Esling, G.~Carpentier, and C.~Agon, ``Dynamic musical orchestration using
  genetic algorithms and a spectro-temporal description of musical
  instruments,'' \emph{Applications of Evolutionary Computation}, pp. 371--380,
  2010.

\bibitem{pachet2016joyful}
F.~Pachet, ``A joyful ode to automatic orchestration,'' \emph{ACM Transactions
  on Intelligent Systems and Technology (TIST)}, vol.~8, no.~2, p.~18, 2016.

\bibitem{handelman2012automatic}
E.~Handelman, A.~Sigler, and D.~Donna, ``Automatic orchestration for automatic
  composition,'' in \emph{1st International Workshop on Musical Metacreation
  (MUME 2012)}, 2012, pp. 43--48.

\bibitem{taylor2006modeling}
G.~W. Taylor, G.~E. Hinton, and S.~T. Roweis, ``Modeling human motion using
  binary latent variables,'' in \emph{Advances in neural information processing
  systems}, 2006, pp. 1345--1352.

\bibitem{DBLP:journals/corr/YaoCVDD15}
\BIBentryALTinterwordspacing
K.~Yao, T.~Cohn, K.~Vylomova, K.~Duh, and C.~Dyer, ``Depth-gated {LSTM},''
  \emph{CoRR}, vol. abs/1508.03790, 2015. [Online]. Available:
  \url{http://arxiv.org/abs/1508.03790}
\BIBentrySTDinterwordspacing

\bibitem{Fischer2012}
\BIBentryALTinterwordspacing
A.~Fischer and C.~Igel, \emph{Progress in Pattern Recognition, Image Analysis,
  Computer Vision, and Applications: 17th Iberoamerican Congress, CIARP 2012,
  Buenos Aires, Argentina, September 3-6, 2012. Proceedings}.\hskip 1em plus
  0.5em minus 0.4em\relax Berlin, Heidelberg: Springer Berlin Heidelberg, 2012,
  ch. An Introduction to Restricted Boltzmann Machines, pp. 14--36. [Online].
  Available: \url{http://dx.doi.org/10.1007/978-3-642-33275-3_2}
\BIBentrySTDinterwordspacing

\bibitem{hinton2010practical}
G.~Hinton, ``A practical guide to training restricted boltzmann machines,''
  \emph{Momentum}, vol.~9, no.~1, p. 926, 2010.

\bibitem{hinton2002training}
G.~E. Hinton, ``Training products of experts by minimizing contrastive
  divergence,'' \emph{Neural computation}, vol.~14, no.~8, pp. 1771--1800,
  2002.

\bibitem{taylor2009factored}
G.~W. Taylor and G.~E. Hinton, ``Factored conditional restricted boltzmann
  machines for modeling motion style,'' in \emph{Proceedings of the 26th annual
  international conference on machine learning}.\hskip 1em plus 0.5em minus
  0.4em\relax ACM, 2009, pp. 1025--1032.

\bibitem{kiros2015skip}
R.~Kiros, Y.~Zhu, R.~R. Salakhutdinov, R.~Zemel, R.~Urtasun, A.~Torralba, and
  S.~Fidler, ``Skip-thought vectors,'' in \emph{Advances in Neural Information
  Processing Systems}, 2015, pp. 3276--3284.

\end{thebibliography}

\end{document}